\documentclass[a4paper, 10pt, conference]{ieeeconf}      % Use this line for a4
\usepackage{FG2024}

\FGfinalcopy % *** Uncomment this line for the final submission

\usepackage{graphicx}
\usepackage{url} 
\usepackage{diagbox}

\IEEEoverridecommandlockouts                              % This command is only
\title{Massively Annotated Datasets for Assessment of Synthetic and Real Data in Face Recognition}

\author{\parbox{16cm}{\centering
    {\large  Pedro C. Neto$^{1,2}$, Rafael M. Mamede$^{1,2}$, Carolina Albuquerque$^{1,2}$, Tiago Gonçalves$^{1,2}$,  \\Ana F. Sequeira$^{1,2}$}\\
   {\normalsize
    $^1$ Faculty of Engineering of the University of Porto, Porto, Portugal\\
    $^2$ Institute for Systems and Computer Engineering, Technology and Science, Porto, Portugal}}% <-this % stops a space
}

\begin{document}

\ifFGfinal
\thispagestyle{empty}
\pagestyle{empty}
\else
\author{\parbox{16cm}{\centering
   {\large  Pedro C. Neto$^{1,2}$, Rafael M. Mamede$^{1,2}$, Carolina Albuquerque$^{1,2}$, Tiago Gonçalves$^{1,2}$,  \\Ana F. Sequeira$^{1,2}$}\\
   {\normalsize
   $^1$ INESC TEC, Porto, Portugal\\
   $^2$ FEUP, Porto, Portugal}}% <-this % stops a space
}
\pagestyle{plain}
\fi
\maketitle

%%%%%%%%%%%%%%%%%%%%%%%%%%%%%%%%%%%%%%%%%%%%%%%%%%%%%%%%%%%%%%%%%%%%%%%%%%%%%%%%
\begin{abstract}

Face recognition applications have grown in parallel with the size of datasets, complexity of deep learning models and computational power. However, while deep learning models evolve to become more capable and computational power keeps increasing, the datasets available are being retracted and removed from public access. Privacy and ethical concerns are relevant topics within these domains. Through generative artificial intelligence, researchers have put efforts into the development of completely synthetic datasets that can be used to train face recognition systems. Nonetheless, the recent advances have not been sufficient to achieve performance comparable to the state-of-the-art models trained on real data. 

To study the drift between the performance of models trained on real and synthetic datasets, we leverage a massive attribute classifier (MAC) to create annotations for four datasets: two real and two synthetic. From these annotations, we conduct studies on the distribution of each attribute within all four datasets. Additionally, we further inspect the differences between real and synthetic datasets on the attribute set.  When comparing through the Kullback–Leibler divergence we have found differences between real and synthetic samples. Interestingly enough, we have verified that while real samples suffice to explain the synthetic distribution, the opposite could not be further from being true. 
\end{abstract}

%%%%%%%%%%%%%%%%%%%%%%%%%%%%%%%%%%%%%%%%%%%%%%%%%%%%%%%%%%%%%%%%%%%%%%%%%%%%%%%%
\section{Introduction}

Complex Face Recognition systems have matched and surpassed human-level performance~\cite{taigman2014deepface}. Recent advances led to deep learning-based neural networks that can learn to distinguish between the most variate identities from a single image. The resourcefulness of such models has led to a continuous focus on improving the best-performing models. Over the years, this improvement was supported by three strong pillars. 1) Exponential increase in computing power; 2) Novel architectures and more expressive deep learning models; 3) Very large datasets.  

As mentioned, one of the approaches to further enhance these models relied on the collection and curation of large datasets~\cite{zhu2021webface260m}. These datasets vary in the number of identities, from 10k to 672k, and in the number of images, from 500k to 17M~\cite{yi2014learning,an2021partial, NechMegaFace2017}. However, the collection of these datasets has raised privacy and ethical concerns regarding the consent of the individuals present in the data~\cite{boutros2023idiff}. This led to the retraction of several of these previously publicly available datasets~\cite{cao2018vggface2}. Moreover, a dataset that is composed of real images with proper curation and consent is not static, since according to the European Union (EU) General Data Protection Regulation (GDPR)~\cite{regulation2016regulation}, consent can be removed at any time. Additionally, individual anonymization of the sample faces is not feasible and further removes the utility of the face to train a face recognition system~\cite{meden2021privacy}.  This poses a problem for current face recognition research, which requires the use of large-scale datasets that are being limited and removed from public access as previously mentioned.

Recently, there has been significant growth in generative artificial intelligence approaches, leading to state-of-the-art methods that can synthesise images that closely resemble real images~\cite{kim2023dcface, alayrac2022flamingo}. Since the initial generative adversarial neural network (GAN)~\cite{goodfellow2014generative} and their improved versions~\cite{karras2018progressive}, there have been several advances that led to the development of diffusion models. These models are easier to train and lead to better-quality images. Recently, some generative models have been proposed for face data, allowing researchers to condition the identity or other attributes~\cite{boutros2023idiff,boutros2024sface2}. Following the improvements in generative artificial intelligence, researchers have redirected their efforts into how to synthesise new datasets for face recognition that could remove the dependency on the previously used real datasets. 

\begin{figure}
    \centering
    \includegraphics[width=0.96\linewidth]{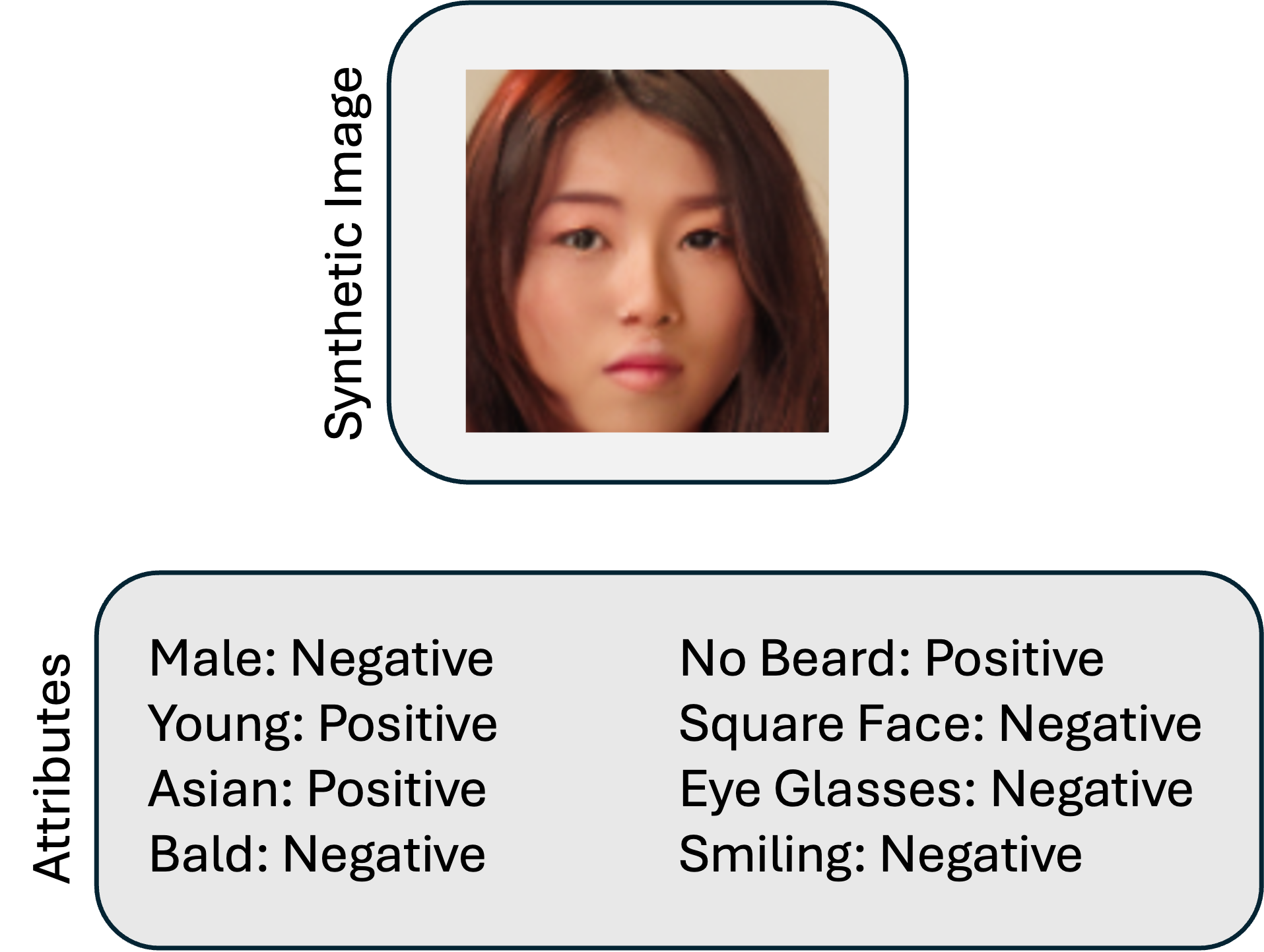}
    \caption{An example of an annotated synthetic image. It is possible to observe some of the well defined attributes.  }
    \label{fig:distribution_illustration}
\end{figure}

Despite efforts to develop synthetic data that faithfully represents real data, the performance of models trained on these new synthetic datasets is yet to achieve a performance similar to models trained on real data~\cite{boutros2023idiff}, even when the data is constructed through different synthesising approaches~\cite{bae2023digiface}. These former models seem to perform considerably worse on certain ethnicities and other variations of the traditional face verification setting, such as cross-pose or cross-age. This behaviour can be sustained by several factors, and Huber~\textit{et al.}~\cite{huber2024bias} has already explored the diversity of synthetic datasets with regards to gender, ethnicity, age and head position. We firmly believe that the diversity of face datasets can be further described by other attributes. The work of Terhörst  \textit{et al.} \cite{terhorst2021maadface} aims to create datasets that are annotated for 47 distinct attributes, which can be leveraged to highlight differences between synthetic and real datasets. Previous research has noted a potential domain gap between real and synthetic data~\cite{DBLP:conf/aaai/XuZNLWTZ20,DBLP:conf/cvpr/Sankaranarayanan18,DBLP:conf/cvpr/LeePYL20}

As seen in Figure~\ref{fig:distribution_illustration}, the performance of an FR system trained on synthetic samples might be restricted by the fact that those samples do not capture the complete variation and full spectrum of real samples. In this paper, we aim to understand how closely the synthetic data mimics the distribution of the real data. For this, we have leveraged two real datasets BUPT-BalancedFace and BUPT-GlobalFace~\cite{wang2021meta} in addition to two synthetic datasets generated, one generated through diffusion (IDiff-Face~\cite{boutros2023idiff}) and the other with a GAN~\cite{boutros2022quantface}. Using Terhörst~\textit{et al.}~\cite{terhorst2021maadface} MAC method, we have computed annotations for all the samples in the four datasets. Following this, we have conducted several studies on the distribution of these annotations in order to extract information regarding the diversity of each dataset.  

This paper is organized into four main sections. The first of these, presents related work regarding automatic annotation strategies and synthetic data generation. Afterwards, in the Methods section, the fundamental details of the experiment's design and setup are discussed in detail. This latter section is followed by a Results section, which aims to present the main findings. Finally, we conclude with a discussion of future work and a summary of the most important elements of this research.  The contributions of this work are the following: 
\begin{itemize}
    \item Created annotations, which will be publicy available, for two real datasets. One of the datasets is balanced for ethnicity, whereas the other follows the world ethnicity distribution;
    \item Replicated the annotation process on two synthetic datasets, enabling future research on the soft-biometrics of these datasets using the released annotations; 
    \item Performed a statistical analysis of the diversity of these datasets through the study of their annotations. 
\end{itemize}

\section{Related Work}

Face recognition from synthetic data has grown in popularity in recent years. It presented some challenges to researchers as it was not possible to generate several samples from the same identity, nor generate sufficiently realistic samples. Over the years, these problems have been mitigated~\cite{deandres2024frcsyn}. In 2022, Boutros~\textit{et al.}~\cite{boutros2022sface} proposed SFace, a generative adversarial network-based approach to create new samples, and in 2024 proposed SFace2~\cite{boutros2024sface2} achieving state-of-the-art results when compared to other GAN-based methods~\cite{kolf2023identity,qiu2021synface}. To mitigate the existence of a single image per identity, some works have also explored unsupervised approaches to train face recognition systems~\cite{boutros2023unsupervised}. IDiff-Face~\cite{boutros2023idiff} comprised a novel diffusion-based technique that conditions the model on a desired identity, leading to a dataset that achieves, in some datasets, a performance comparable to a model trained on real datasets. Kim~\textit{et al.}~\cite{kim2023dcface} leveraged the conditioning of diffusion models to generate samples from a specific identity and with a specific style. For instance, it is possible to generate an image of a certain person using glasses.  

Soft-biometric annotations for face images provide contextual information not dependent on a specific identity (such as gender, age, or ethnicity) and are essential for exploring the variability of the data. Low variability of some characteristics can make models trained on that dataset less robust to applications on real-world data, where inference on examples with such characteristics is necessary.
With this information, it is also possible to explore, disclose, and correct demographic biases, addressing fairness concerns. The process of annotating manually is labour-intensive, which can be unfeasible for large datasets. Some works have proposed classification-based estimation of soft-biometric characteristics which can aid the annotation of large datasets: Karkkainen  \textit{et al. } \cite{karkkainen2021fairface} (ethnicity, gender, age), Gonzalez-Sosa \textit{et al. } \cite{Gonzalez2018Facial} (gender, age, craniofacial features, skin colour, subjective annotation),Mirabet~\textit{et al.}~\cite{mirabet2023lvt} (hidden face attributes), Merler \textit{et al. } \cite{merler2019diversity} (gender, age, glasses, beard, and moustache), Terhörst  \textit{et al.} \cite{terhorst2021maadface} (47 attribute annotations covering gender, ethnicity, age, accessories, facial hair, hairstyles, subjective annotation, and other facial characteristics). Following the completeness of the 47 attributes generated by the model proposed by Terhörst  \textit{et al.}, it presents itself as the ideal approach to studying the broad diversity of synthetic datasets. A similar annotation approach has been followed for DeepFake datasets, and through the analysis of such annotations it was possible to understand and detect certain biases on DeepFake detection systems~\cite{xu2022comprehensive}. This further highlights the importance of having this information available.

\section{Methods}

\subsection{Datasets}

In this section we present the four datasets used in our experiments. We provide details on their composition and, in the case of synthetic datasets, the generative approach. 

\subsubsection{Real datasets}

BUPT-Balanced and BUPT-GlobalFace have been proposed by Wang~\textit{et al.}~\cite{wang2021meta} and were intended to create a framework to study the biases of face recognition models. Each identity on the dataset has been labelled according to its skin tone into one of the following ethnicities: African, Asian, Caucasian and Indian. BUPT-Balanced balances the number of identities that belong to each of these four categories and is composed of 1.3 million images with 28k identities, which means that there are 7k identities per ethnicity. On the other hand, BUPT-Globalface contains two million images from 38k identities, and the ethnicity distribution of the identities follows the same distribution seen in the world's population. 

Comprising two distinct ethnicity distributions these datasets allow for a comparison of the Real vs. Real to be used as a baseline of the differences that can be expected from datasets that are known to have different distributions on certain attributes. 

\subsubsection{Synthetic datasets}

For the study of synthetic data, we have selected two fundamentally different datasets. The first dataset, referred to throughout the rest of this paper as Syn-GAN, was introduced as a tool to be used in quantisation scenarios~\cite{boutros2022quantface}. It is comprised of 500k images that have no information regarding their identity. They were generated with a generative adversarial network~\cite{goodfellow2014generative}. Using noise sampled from a Gaussian distribution, a pretrained generator\footnote{\url{https://github.com/NVlabs/stylegan2-ada}} of StyleGAN2-AD is used to create novel face samples. Neto~\textit{et al.}~\cite{neto2023compressed} explored the effect of the quantisation of deep neural networks with this synthetic dataset on the bias of the final face recognition system. The higher robustness shown highlights a possibility that this data comprises samples that slightly deviate from the real data distribution. 

The second dataset exploited diffusion models to generate identity-conditioned samples. IDiff.Face~\cite{boutros2023idiff} was used to create a dataset called CPD-25 (Two-Stage), which comprises 10k identities and 50 images for each identity. This dataset is significantly more realistic and has been shown to have a performance that reduces the gap between models trained on real and synthetic datasets. Besides the identity, no other attribute is conditioned. 

\subsubsection{Annotations}

None of the datasets includes annotations beyond identity, with an exception for the skin tone-based labels on the two real datasets. Hence, studies on the diversity of these datasets from the point of view of soft-biometrics was limited. Additionally, fairness assessments were not trivial. Knowing the impact that these annotations might have no future research, we released, for each image in each dataset, 45 different attributes\footnote{\url{https://github.com/NetoPedro/Synth-MAAD-Face}}. For ethical reasons, we have decided to exclude annotations for the fields ``Chubby" and ``Attractive" seen in the original paper. In total we provide roughly 189M annotations, which is slightly larger than the annotations given by Terhörst  \textit{et al.}~\cite{terhorst2021maadface} at 124M.

\subsection{Experimental Design}

\subsubsection{Annotation process}

The generation of annotations for each of the aforementioned datasets used the methodology proposed by Terhörst  \textit{et al.} \cite{terhörst2020beyond, terhorst2021maadface}. Given an image $x$ aligned with MTCNN~\cite{zhang2016joint}, we map the image to face template space given by FaceNet~\cite{schroff2015facenet}, which is further mapped to 47 different soft-biometric attributes via the Massive Attribute Classifier (MAC), a multi-objective NN-based classifier. Each attribute is considered to be an individual classification task, and the majority of the attributes can be ``Positive", ``Negative" or ``Undefined". 

The proposed implementation strategy leveraged the idea of reliability as a metric for the confidence of a prediction of each attribute \cite{terhorst2019reliable}. Here, we consider an ensemble of $m$ MAC classifiers where dropout is applied individually with a given probability ($p_{drop}=0.5$) during test. This results in $m$ estimators with slightly different architectures due to the zeroed connections. This dropout process mimics the behaviour of a Gaussian process and the final prediction for each of the $A$ attributes is computed through a majority vote of the classifiers in the ensemble. Considering each element of the ensemble as a classifier $f_{i}:R^N \rightarrow [0,1]^A$, where $i\in\{1,2,...,m\}$, the reliability of the prediction for the attribute $a$,  $rel(f(x)^{(a)})=rel(y^{(a)})$, for $a\in\{1,2,...,A\}$, is given by:

\begin{equation}
    rel(y^{(a)}) = \frac{(1- \alpha)}{m} \sum_{i=1}^{m} y_{i}^{(a)} - \frac{\alpha}{m^2} \sum_{i=1}^{m}  \sum_{j=1}^{m} |y_{i}^{(a)}-y_{j}^{(a)}|
    \label{Rel}
\end{equation}

The parameter $\alpha$ balances the impact of the centrality measure versus the impact of the dispersion metric and, for equal impact of these factors, we chose this value to be 0.5. Similarly to the original paper, the number of estimators was set as $m=100$.

The original implementation of MAC creates a new ensemble of models for each image, and that ensemble is then used to estimate the soft-biometric attributes. Hence, each image is passed, one by one, 100 times through the network. Our implementation creates a batch of images during the inference step. Each image in the batch will be evaluated by the same ensemble of models. However, since the same image is still seen by 100 distinct architectures, we retain the advantages of the Gaussian Process while being able to run experiments hundreds of times faster.  This allows the usage of the MAAD methodology to be more easily applicable to label large datasets. We used a labelling batch size of 1024.

\begin{figure*}
    \centering
    \includegraphics[width=1\linewidth]{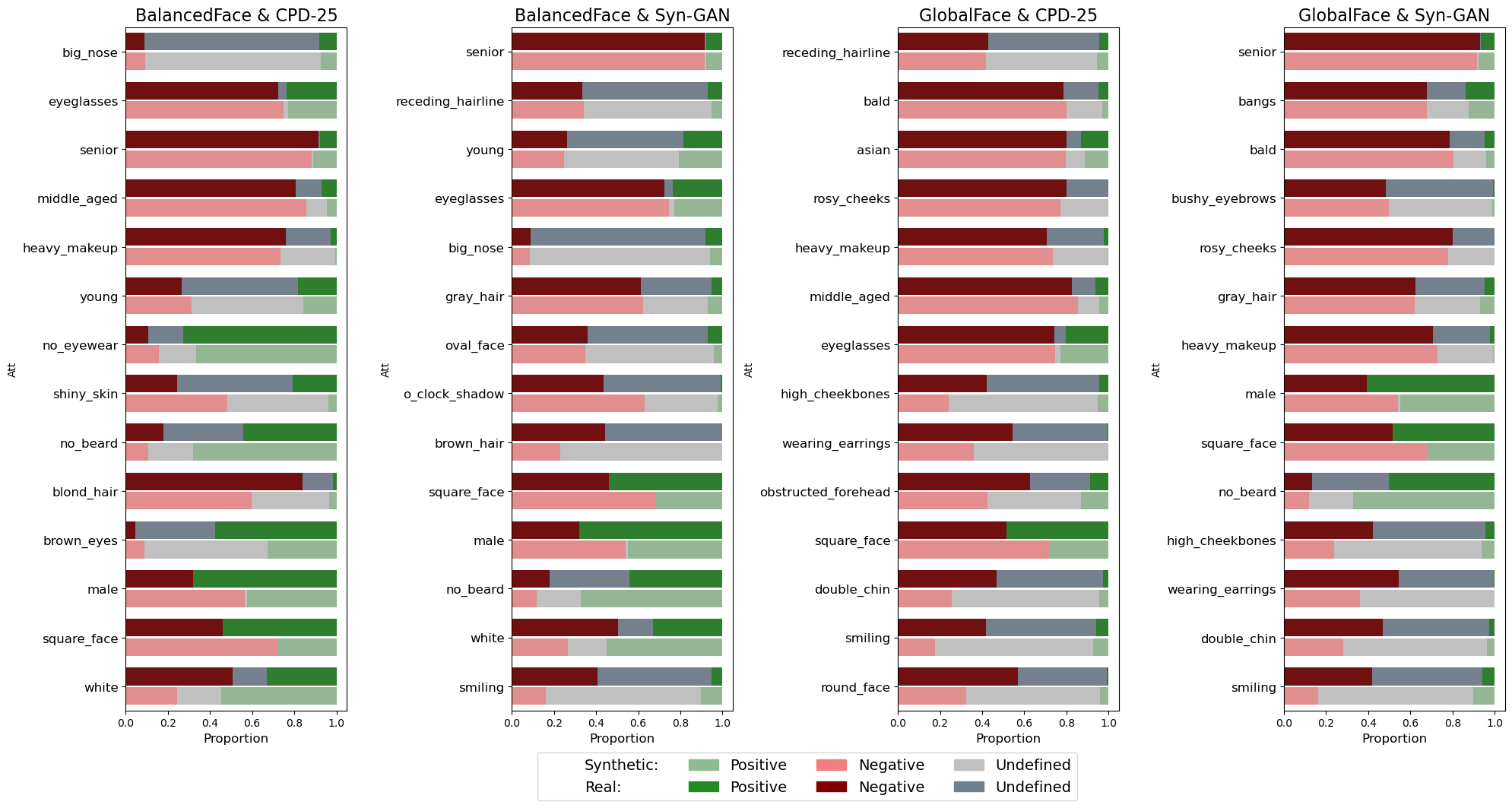}
    \caption{Comparison of the real and the synthetic datasets on individual attribute distribution. From the left to the right we have a comparison between BalancedFace and CPD-25, BalancedFace and Syn-GAN, GlobalFace and CPD-25, and GlobalFace and Syn-GAN. The first seven entries of each plot represent the most similar attribute distributions, whereas the seven bottom attributes represent the less similar distributions.  }
    \label{fig:attribute_dist}
\end{figure*}

\subsubsection{Comparison of the different datasets}

Comparing two datasets at the image level is a non-trivial task. Hence, dataset comparison is easier if done at the level of the latent space of the deep neural network. However, with regards to faces, it might not encapsulate the variety of attributes that might be present in an image, since two individuals with a very similar set of attributes will have some distance between them. Additionally, it is not direct to understand the ``differences" in this space. However, in the MAAD attribute space, not only comparing different samples with clarity is easier, but individuals with similar characteristics share the space.

Taking into consideration the advantages of this attribute-space, we have devised several strategies to measure the discrepancy between real and synthetic data. One of the first approaches was to measure the relative frequency of ``Undefined" predictions on each of the four datasets, which indicated very similar results on average. Afterwards, we focused on the comparison of individual attributes and how much different was the prevalence of positives and negatives for datasets of different sources. This brought some interesting perspectives on the distribution of each attribute. 

Additionally, and considering that learnt models might be useful to detect the distinct patterns of data sources, we attempted to measure the relative classifiability of ``Real vs. Synthetic". This was done with two strategies: using a classifier; creating two clusters with K-Means~\cite{macqueen1967some} and validating how many samples of each source fall within each of the learnt clusters.

Considering the fact that the individual analysis of the distribution of a single dataset is a poor metric, we propose to model the prediction of each dataset with a Kernel Density Estimation~\cite{davis2011remarks, parzen1962estimation} approach on the attribute space. This information takes into consideration the several configurations that each individual might take. Before computing the distribution we take the mode of all the attribute-sets of an identity. Finally, having learnt the distribution of each dataset, we can compute the Kullback–Leibler divergence (Eq.~\ref{KL})~\cite{kullback1951information}, on both sides, between real and synthetic datasets. 

\begin{equation}
    D_{KL}(P||Q) = \sum_{x\in X} P(x) log(\frac{P(x)}{Q(x)})
    \label{KL}
\end{equation}

This distance function has the particularity of not being a metric, since it does not respect the symmetry property. Hence results differ if we swap $P$ and $Q$. A possible interpretation of this distance is the quantification of the information lost when using $Q$ to approximate the distribution $P$. This is particularly useful in our scenario as it can tell us the information lost when we approximate the distribution, at the attribute space, of real data using synthetic data. Ideally, this distance should be close to zero as the information contained in one distribution would be reflected in the other distribution.

\subsection{Experimental Setup}

All the annotation experiments were conducted in a GPU cluster, leveraging a NVIDIA A100 GPU with 80GB of VRAM. The batch size for the inference was set at 1024, and we have conducted several tests to find the optimal batch size for our configuration. Additionally, we have measured the impact of having different batch sizes on the predictions and did not find statistically significant differences between different runs. We separated the face template extraction and the attribute computation steps so that it could be possible to create batches on the latter stage without affecting the approximation to a Gaussian Process. 

The remaining experiments were conducted in a consumer grade laptop without GPU. Annotations were saved for later use and release.

\section{Results}

\subsubsection{A study on the attributes}

Following the previously described methodology, we aimed to understand how the different datasets behaved with respect to their attributes. In the initial stage, we aimed to understand the attributes of each dataset individually, hence, for the different combinations of real datasets with synthetic datasets we calculated the seven most similar attribute distributions and the seven most non-similar. 

\begin{figure}
    \centering
    \includegraphics[width=0.85\linewidth]{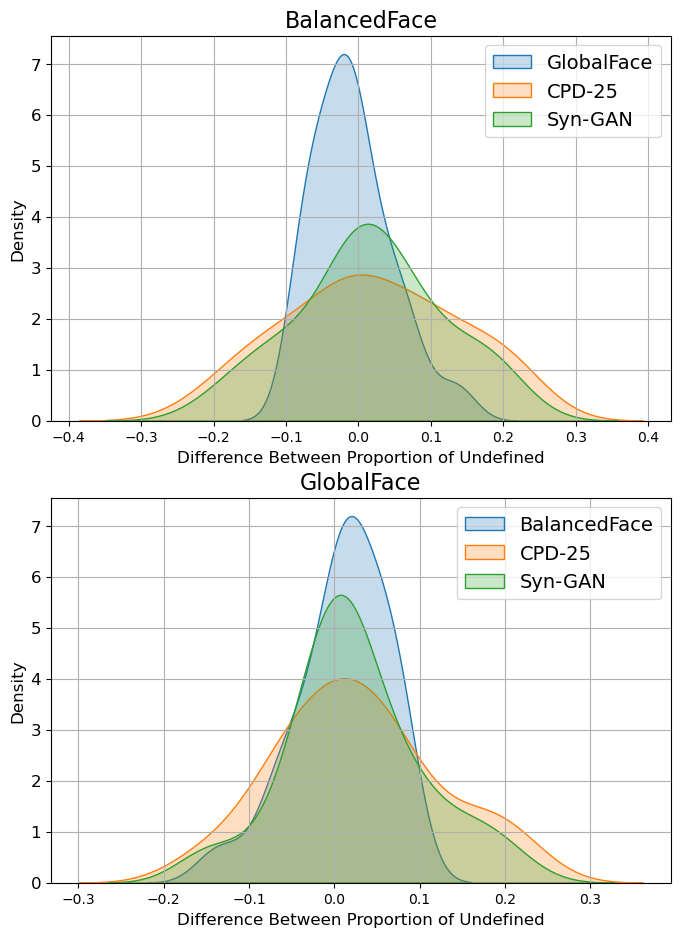}
    \caption{Distributions $Prop0_{d1}^{(a)}-Prop0_{d2}^{(a)}$, where $Prop0_{d'}^{(a')}$ represents the proportion of the Undefined prediction of MAC for attribute $a'$ in dataset $d'$. The title of each plot represents the $d2$ set, and the label of each KDE plot represents the $d1$ set.}
    \label{fig:zero_prop_diff}
\end{figure}
Looking at Figure~\ref{fig:attribute_dist} it is possible to observe the four different combinations. For each plot, the top seven attributes are the ones that are more in line when the two datasets are compared, whereas the bottom seven are the most dissimilar. Some attributes are frequently displayed in the dissimilar set: Square Face, Male, No Beard and Smiling. In the synthetic data, samples are generally more skewed towards being female, whereas in real data the opposite happens. Probably related to this prevalence, there are significantly fewer samples with beards on synthetic datasets. Although it seems a minor issue, the beard represents one of the most natural and common forms of face occlusions. One other relevant aspect is the inability to properly detect smiles on synthetic data, as the majority of the annotations are ``Undefined". This might impact the variability of the samples as the emotions/face reactions are one of the elements that most affect the perception of a face. One additional relevant element is the presence of White as one of the most dissimilar attribute distributions when the real dataset is BalancedFace. While this dataset focuses on balancing the different skin tones, synthetic datasets contain mostly white individuals. 

On the other hand, a few attributes are consistently displayed as ones with the most similar distribution. Age-related attributes such as Senior, Young and Middle-aged are considerably similar in both types of datasets. Although this seems to be a good indicator, we have not measured if the age distribution within each identity is similar. For instance, we have 100 images uniformly spread between 18 and 70 years old, but having only one age group within each identity. Or we could have images of different ages within each identity. It is also visible that on synthetic datasets it is extremely rare to have a sample face wearing heavy makeup.

\subsubsection{Dispersion of Undefined Samples}

Another relevant question we can raise when discussing the labeling of the synthetic data with MAC is whether we get an abnormal amount of Undefined predictions. Abnormal behavior of this outcome could indicate underlying problems of the labeling method, and its adequacy when applied to our synthetic datasets. If we noticed that in general there is a higher amount of this label across all attributes in the synthetic datasets, we could infer that the MAC is having trouble defining meaningful predictions for each class, either because the synthetic data has a poor representation of that attribute, or it is not represented.  

To verify underlying trends in the Undefined prediction, we analyse the differences between the proportions of this label, given two datasets, for each attribute. We take notice of changes in the behavior of these distributions when analysing Real-Real or Synthetic-Real differences. The results, Figure \ref{fig:zero_prop_diff}, show no significant difference in the mean of these differences, meaning we do not observe generalised trends for Undefined labeling across all attributes. What we can verify is the difference in the dispersion of these differences, meaning that some attributes have more extreme fluctuations. We also noted that the attributes that have the highest decrease in the proportion of Undefined when compared to a Real dataset were consistently attributes related to facial hair (such as 'o\_clock\_shadow', 'no\_beard', 'sideburns', 'goatee' ). This is consistent with a balancing of the gender attribute in the synthetic dataset, since prediction for these attributes tends to be easier in the 'female' class.   On the other hand attributes that consistently have an increase in unpredictability include 'smiling' and accessory use such as 'wearing lipstick' and 'wearing earrings'. Emotions and their expressions might be particularly difficult to model with a generative system, as well as artifacts such as accessories. It might be the case that since these share no link to the identity information, they might trigger the generative model to ignore or smooth them.

\begin{figure}
    \centering
    \includegraphics[width=1.05\linewidth]{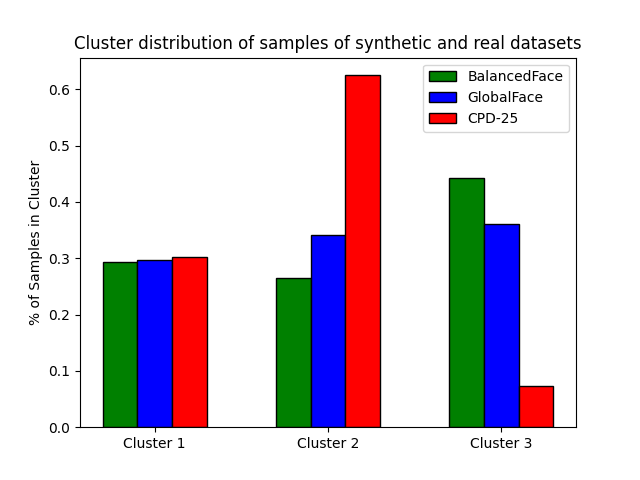}
    \caption{Cluster aggregation of the different identities on each of three datasets (BalancedFace, GlobalFace and CPD-25). Clusters calculated with K-Means. Syn-GAN was removed from the comparison, since it consists of 500k images of distinct identities.  Synthetic data has higher presence in cluster 2 ($>60\%$) and lower in cluster 3 ($<10\%$).}
    \label{fig:cluster_datasets}
\end{figure}

\subsubsection{Clustering Synthetic vs. Real}

In addition to the comparison of the different attributes with respect to their real and synthetic distribution, we have attempted to fit a k-means ($k=3$) model to three out of the four datasets. Figure~\ref{fig:cluster_datasets} shows the distribution of the samples of each dataset within one of the three clusters. As expected, BalancedFace and GlobalFace share very similar distributions across clusters. Surprisingly, the presence of CPD-25 in the third cluster is rather small, and the dataset is heavily inserted in the second cluster. This evidence already highlights some potential differences between the attribute space of synthetic and real samples. Especially if we consider that even on the most prevalent cluster for synthetic data, real data is significantly present, whereas the opposite is not verifiable.

\begin{table}[!ht]
    \centering
    \caption{KL divergence between all the distribution learnt by KDE for all four datasets. Interesting to denote the lower information loss when real data is used to approximate synthetic data in comparison with the higher information loss when we swap P and Q. }
    \begin{tabular}{|c|c|c|c|c|}
    \hline
        \diagbox{P}{Q} & GlobalFace & BalancedFace & Syn-GAN & CPD-25  \\
    \hline
         GlobalFace & - & 0.708& 3.007 & 2.223\\
         BalancedFace & 0.409 & - & 1.984 & 1.223\\
         Syn-GAN & 1.010 & 0.613 & - & 0.249\\
         CPD-25 & 0.507 & 0.247 & 0.371 & -\\
    \hline
    \end{tabular}
    
    \label{tab:table_KL}
\end{table}

\subsubsection{Synthetic vs. Real Divergence}

Following the potential difference in the distribution of the attribute set of real and synthetic datasets, we have used a KDE model to estimate and approximate the distribution for each one of the four datasets. Afterwards, as seen in Table~\ref{tab:table_KL}, we have measured the Kullback–Leibler divergence for all combinations of datasets. When P is set to the distribution of GlobalFace, we noticed that both Syn-GAN and CPD-25 as Q lead to significant information loss. On the other hand, BalancedFace as Q allows for diminished information loss. Setting BalancedFace as P leads to very similar results with a synthetic Q, but considerably better than the previous comparison. On the other hand, GlobalFace is quite accurate at approximating the other real datasets. Both synthetic datasets can be easily explained by the other datasets, leading to very small values of information loss.  This discrepancy in the KL values highlights the lack of diversity shown in synthetic datasets for face recognition and how they must improve to replace real data.  

\section{Conclusion}

Throughout this paper we have covered several strategies to uncover the reason behind the inferior performance of models trained on synthetic data when compared to models trained on real data. Acknowledging the difficulties of comparing these datasets at image level, we have proposed to use MAC to create massive annotations for each of four datasets. This process allowed for a few studies on the diversity and gap between real and synthetic datasets. 

Leveraging the annotations, it was possible to immediately inspect some attributes and their difference in distribution across all datasets. It was also possible to measure the undefined dispersion on synthetic datasets and uncovering that attributes such as smiling are difficult to measure on synthetic data, as shown by the quantity of undefined samples for this attribute. 

Considering the attribute set as a whole combination of attributes, it was possible to place the samples of each dataset on one of two cluster and extract hints regarding the lower variability of synthetic data. Additionally, after modelling this distributions, we manage to user the Kullback–Leibler divergence to measure the information difference between the four datasets. As expected, synthetic datasets shown a poor capability to approximate real data.

In summary, we have not yet find a clear answer to the reason behind the performance difference of models trained on these datasets. Yet, we have made contributions on the gaps between both types of datasets and we have released the annotations. Future researchers can leverage these annotations to further condition diffusion models, to find correlations between a set of attributes and the performance, or to build better automatic annotation tools. There are several directions in which this research might lead.

% Acknowledgments
\section{Acknowledgments}
This work was financed by National Funds through the Portuguese funding agency, FCT - Fundação para a Ciência e a Tecnologia, within the PhD grants with the references ``2020.06434.BD'' and ``2021.06872.BD'', and within project UIDB/50014/2020.
DOI 10.54499/UIDB/50014/2020 | https://doi.org/10.54499/uidb/50014/2020.

% Bibliography
{\small
\bibliographystyle{ieee}
\bibliography{egbib}
}

\end{document}